\documentclass[conference]{IEEEtran}
\IEEEoverridecommandlockouts
\usepackage{cite}
\usepackage{amsmath,amssymb,amsfonts}
\usepackage{algorithmic}
\usepackage{graphicx}
\usepackage{textcomp}
\usepackage{xcolor}
\def\BibTeX{{\rm B\kern-.05em{\sc i\kern-.025em b}\kern-.08em
    T\kern-.1667em\lower.7ex\hbox{E}\kern-.125emX}}

\usepackage{amsmath}
\usepackage{booktabs}
\usepackage{multirow}
\usepackage{algorithm}
\usepackage{algorithmic}

\begin{document}

\title{LBPE: Long-token-first Tokenization to Improve Large Language Models}

\author{\IEEEauthorblockN{Haoran Lian}
\IEEEauthorblockA{\textit{Beihang University} \\
Beijing, China \\
lianhaoran@buaa.edu.cn}
\and
\IEEEauthorblockN{Yizhe Xiong}
\IEEEauthorblockA{\textit{Tsinghua University;BNRist} \\
Beijing, China \\
xiongyizhe2001@163.com}
\and
\IEEEauthorblockN{Zijia Lin}
\IEEEauthorblockA{\textit{Tsinghua University;BNRist} \\
Beijing, China \\
linzijia07@tsinghua.org.cn}
\and
\IEEEauthorblockN{Jianwei Niu}
\thanks{Jianwei Niu is corresponding author.}
\IEEEauthorblockA{\textit{Beihang University} \\
Beijing, China \\
niujianwei@buaa.edu.cn}
\and
\IEEEauthorblockN{Shasha Mo}
\IEEEauthorblockA{\textit{Beihang University} \\
Beijing, China \\
moshasha@buaa.edu.cn}
\and
\IEEEauthorblockN{Hui Chen}
\IEEEauthorblockA{\textit{Tsinghua University;BNRist} \\
Beijing, China \\
jichenhui2012@gmail.com}
\and
\IEEEauthorblockN{Peng Liu}
\IEEEauthorblockA{\textit{Beijing Institute of Technology} \\
Beijing, China \\
773798069@qq.com}
\and
\IEEEauthorblockN{Guiguang Ding}
\IEEEauthorblockA{\textit{Tsinghua University;BNRist} \\
Beijing, China \\
dinggg@tsinghua.edu.cn}
}

\maketitle

\begin{abstract}

The prevalent use of Byte Pair Encoding (BPE) in Large Language Models (LLMs) facilitates robust handling of subword units and avoids issues of out-of-vocabulary words. Despite its success, a critical challenge persists: long tokens, rich in semantic information, have fewer occurrences in tokenized datasets compared to short tokens, which can result in imbalanced learning issue across different tokens. To address that, we propose LBPE, which prioritizes long tokens during the encoding process. LBPE generates tokens according to their \texttt{reverse ranks of token length} rather than their ranks in the vocabulary, granting longer tokens higher priority during the encoding process. Consequently, LBPE smooths the frequency differences between short and long tokens, and thus mitigates the learning imbalance. Extensive experiments across diverse language modeling tasks demonstrate that LBPE consistently outperforms the original BPE, well demonstrating its effectiveness.
\end{abstract}


\section{Introduction}
Recently, Large Language Models (LLMs) have become a burgeoning paradigm in handling a broad array of Natural Language Processing (NLP) tasks. The tokenization process in most modern LLMs \cite{radford2019language,brown2020language,rae2021scaling,zhang2022opt,biderman2023pythia,touvron2023llama,yang2023baichuan,achiam2023gpt,dubey2024llama} employs Byte Pair Encoding (BPE) \cite{sennrich2015neural}, a method that was originally designed for data compression \cite{gage1994new}. The adoption of BPE in LLMs is driven by its capability to decompose words into smaller, manageable subword units, thus avoiding out-of-vocabulary words, facilitating flexible and semantically accurate representations of input data. 

BPE consists of two main stages. In the training stage, BPE iteratively merges the most frequent pair of existing tokens (which are initialized with unit tokens like bytes or characters) in a corpus into a new token, and adds it to the vocabulary until a desired vocabulary size is reached. In the encoding stage, following the ranks of tokens in the vocabulary as merging priority (i.e., tokens added earlier have higher frequency and thus are assigned higher priority to be merged into), token pairs are iteratively merged to build the token representation for a given text. 
Since its inception, BPE has undergone various modifications to better suit the needs of complex NLP tasks, including identifying the optimal vocabulary size for various tasks \cite{xu2020vocabulary,gutierrez2021characters}, removing redundant tokens in the vocabulary \cite{lian2024scaffold}, etc.

However, existing studies for BPE have overlooked a challenge in language model training: while the long tokens usually have complex semantic information, they have relatively fewer occurrences in the tokenized dataset compared to the short tokens. Specifically, the original BPE merges token pairs according to their ranks in the vocabulary. And short tokens usually have high ranks due to their high frequency. Hence BPE intends to merge the short tokens first, leading to a lower frequency of long tokens. Such disparity in token frequencies can result in imbalanced learning difficulties across different tokens. Long tokens, due to their lower individual occurrence frequencies, are notably harder to learn for models \cite{su2024mile,su2024maskmoe}.

\begin{figure}[t]
\centerline{\includegraphics[width=0.9\columnwidth]{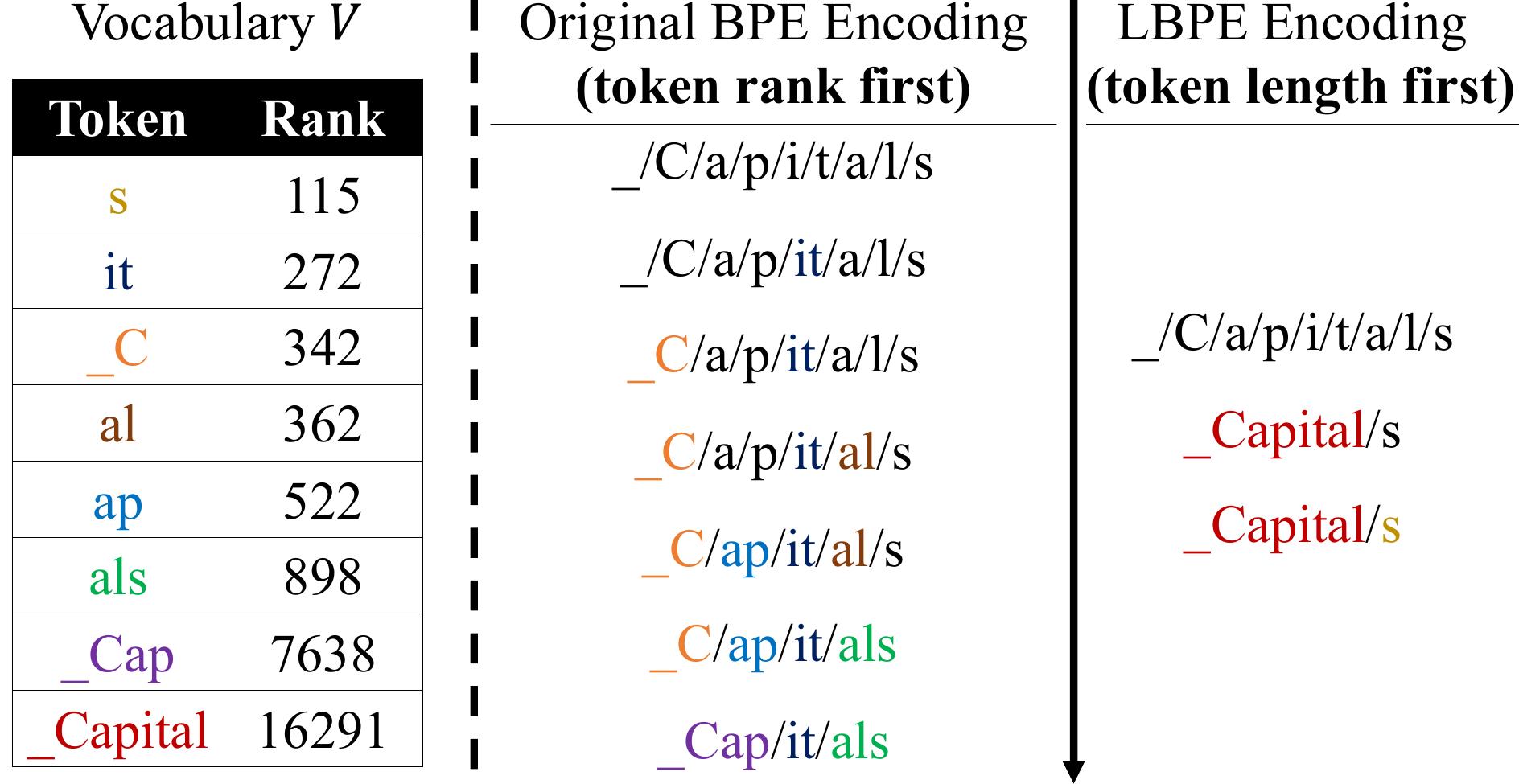}}
\caption{Compared to the original BPE that iteratively merges token pairs according to the rank in vocabulary, LBPE merges a sequence of unit tokens into the final token representation directly according to the \textbf{\texttt{reverse rank of token length}}. Then LBPE can encode ``\_Capital" as one token which is semantically more accurate. However, the original BPE merges ``al" and ``s" first according to the rank in vocabulary, resulting in the failure to derive ``\_Capital" in the end. ``\_" represents space character.}
\label{encoding}
\end{figure}

To address that issue, we propose enhancements to the BPE encoding algorithm, aiming to mitigate the imbalanced learning issue between short tokens and long tokens. Specifically, we propose the simple and effective \textbf{LBPE} with a long-token-first encoding algorithm, which is parameter-free, computation-light, easy-to-implement, and widely effective. 
As shown in Fig. \ref{encoding}, in the encoding stage, the proposed LBPE does not merge token pairs iteratively, but merges a sequence of unit tokens into the final token representation directly. Notably, the merging priority is according to the \textbf{\texttt{reverse rank of token length}} (i.e., the longer tokens have higher priority for merging), rather than the rank of token in the vocabulary. 
Thanks to such modifications, LBPE can help long tokens have more occurrences in the final token representations of a given text dataset, thus making the frequency of tokens with different lengths more balanced.

We conduct extensive experiments on language modeling tasks. Results on 7 widely used benchmarks demonstrate that LBPE consistently outperforms the original BPE for pretraining LLMs from scratch. Particularly, we further show that LBPE is substantially useful in continual pretraining for an existing LLM, so it's never too late to use LBPE. Furthermore, LBPE is orthogonal to existing modifications on BPE, like Scaffold-BPE \cite{lian2024scaffold}, and can be combined with them to achieve further improvements.

Overall, our contributions are three-fold:
\begin{itemize}
    \item We observe that the rank-first iterative encoding process of BPE intends to merge natively high-frequency short tokens first, which exacerbates the imbalanced learning issue across different tokens.
    \item We propose LBPE, which merges a sequence of unit tokens directly according to their reverse ranks of token length in the encoding stage. LBPE is parameter-free, computation-light, easy-to-implement, and widely effective.
    \item Extensive experiments demonstrate that LBPE outperforms the original BPE on diverse language modeling tasks, proving its effectiveness and robustness.
\end{itemize}

\section{Methodology}

The encoding process of the original BPE encodes a text $T$ into a token representation (i.e., $R$) using the vocabulary $V$ generated by BPE training. Firstly, $R$ is a sequence of the smallest unit tokens (i.e., character/byte tokens) obtained by splitting $T$, which is denoted as $R^0$. And then, following the ranks of tokens in $V$ as merging priority (i.e., tokens added earlier are assigned higher priority to be merged into), $R$ is iteratively updated, via replacing the top ranked token pair of two adjacent tokens, i.e., $(R_i, R_{i+1})$, as a new token $t \in V$, as formulated below. 
\begin{equation}
t = \arg\min_{t' = (R_i, R_{i+1}) \in V} rank_V(t')
\end{equation}
where $rank_V(t')$ denotes the rank of $t'$ in $V$.

However, the rank-first encoding algorithm intends to have short tokens in the final $R$, which are natively of high frequency and would be learned sufficiently by a model. As for long tokens, their lower frequencies can lead to imbalanced learning for the model. To mitigate that, we propose LBPE, which employs a long-token-first encoding algorithm. Specifically, LBPE uses the \textbf{\texttt{reverse ranks of token length}} as merging priority (i.e., the longer tokens have higher priority for merging), to generate tokens iteratively. In each iteration, the longest unit token span (i.e., a few consecutive unit tokens) in $R^0$ that is not merged in previous iterations would be merged into a new token $t$, as formulated blow:
\begin{equation}
\label{eq:length}
t = \arg\max_{t' = (R^0_i,\ldots, R^0_{i+k}) \in V} k
\end{equation}
where $t' = (R^0_i,\ldots, R^0_{i+k})$ is a unit token span in $R^0$ with a length of $k$. Note that LBPE only modifies the encoding stage of the original BPE and does not modify the training stage. 

Considering that $k$ is always no larger than the maximal token length $m$ of the vocabulary $V$ (i.e., $m = \max_{\forall token \in V} |token|$), we can leverage sliding window algorithm to implement LBPE easily in a greedy manner, as illustrated in Algorithm \ref{Encoding Algorithm of Byte-Pair Encoding}. Specifically, we use sliding windows of lengths decreasing from $m$ to $1$, to search for the longest unit token spans in $R^0$ that can form a token in $V$ and is never merged in previous iterations. By indexing the vocabulary $V$ in a hash map, the searching process can be well accelerated.
The time complexity of the original BPE encoding algorithm and LBPE encoding algorithm is $O(|T|^2)$, $O(m|T|)$, respectively. As usually $m < |T|$, LBPE can have higher encoding efficiency.

\begin{algorithm}[t]
\caption{LBPE Encoding Algorithm}
\label{Encoding Algorithm of Byte-Pair Encoding}
\begin{algorithmic}[1]
\REQUIRE A Text $T$, A Vocabulary $V$
\STATE Split $T$ into a unit token representation list $R^0$
\STATE Initialize window length $l \leftarrow \max_{\forall token \in V} |token|$
\WHILE{$l > 0$}
    \FOR{$i \leftarrow 0$ to $|R^0| - l$}
        \STATE $span \leftarrow R^0[i:i+l]$
        \IF{$span$ in $V$ and $R^0[x]_{\forall x \in [i:i+l)}$ is not marked}
            \STATE Token $t \leftarrow V[span]$
            \STATE Mark $R^0[i:i+l]$ as $t$
        \ENDIF
    \ENDFOR
    \STATE $l \leftarrow l - 1$
\ENDWHILE
\STATE Obtain the final token representation $R$ of $R^0$ in order based on markings
\RETURN $R$
\end{algorithmic}
\end{algorithm}

\section{Experiments}

\begin{table*}[t]
\centering
\caption{At varying model scales, the average 0/5-shot accuracy on LLM benchmarks ($p$-value $<0.01$).}
\begin{tabular}
{cc|ccccccccccc}
\toprule
& & \textbf{BoolQ} & \textbf{HellaSwag} & \textbf{OpenBookQA} & \textbf{PIQA} & \textbf{SIQA} & \textbf{StoryCloze} & \textbf{Winogrande} \\
\midrule
\multirow{2}{*}{468M} & Original BPE & 58.64 & 40.78 & 30.50 & 66.57 & 43.40 & 62.77 & 53.00 \\
& LBPE & \underline{\textbf{60.89}} & \underline{\textbf{41.61}} & \underline{\textbf{31.70}} & \underline{\textbf{67.08}} & \underline{\textbf{43.71}} & \underline{\textbf{63.52}} & \underline{\textbf{54.81}} \\
\midrule
\multirow{2}{*}{1.2B} & Original BPE & 60.86 & 47.25 & 31.70 & 68.55 & 44.09 & 65.61 & 55.52 \\
& LBPE & \underline{\textbf{61.70}} & \underline{\textbf{48.03}} & \underline{\textbf{32.90}} & \underline{\textbf{69.40}} & \underline{\textbf{45.01}} & \underline{\textbf{66.68}} & \underline{\textbf{56.43}} \\
\midrule
\multirow{2}{*}{6.7B} & Original BPE & 62.87 & 60.57 & 35.10 & 73.69 & 46.98 & 71.43 & 60.97 \\
& LBPE & \underline{\textbf{64.10}} & \underline{\textbf{61.60}} & \underline{\textbf{36.70}} & \underline{\textbf{74.32}} & \underline{\textbf{47.93}} & \underline{\textbf{72.88}} & \underline{\textbf{62.31}} \\
\bottomrule
\end{tabular}
\label{Language Modeling Results}
\end{table*}

We employ the recently well-attended language modeling tasks to validate the effectiveness of the LBPE.

\subsection{Experimental Setup}

\label{Experimental Setup}

\subsubsection{Datasets}
Our models are trained on the Pile \cite{gao2020pile} dataset, an 825.18 GiB English text dataset designed for training LLMs. 
The data distribution for our model training is identical to that of the original work \cite{gao2020pile}.

\subsubsection{Tokenizer}
Following LLaMA \cite{touvron2023llama,touvron2023llama2}, we train a 32K character-level vocabulary and tokenize the dataset using the original BPE and LBPE, respectively. 
Similar to GPT-2 \cite{radford2019language}, pre-tokenization was employed to prevent the merging of tokens from different character categories. And following LLaMA \cite{touvron2023llama}, we split numbers into individual digits.

\begin{table}[t]
\centering
\caption{Token Length Distribution on Pile Dataset.}
\label{Token Length Distribution}
\resizebox{\linewidth}{!}{
\begin{tabular}{c|ccccccc}
\toprule
\textbf{Token Length} & \textbf{Original BPE Frequency} & \textbf{LBPE Frequency} \\
\midrule
1-3 & $2.287\times 10^{11}$ & $2.264\times 10^{11}$ (-0.97\%) \\
4-6 & $1.002\times 10^{11}$ & $1.005\times 10^{11}$ (+0.37\%) \\
7-9 & $3.618\times 10^{10}$ & $\mathbf{3.704\times 10^{10}}$ \textbf{(+2.37\%)} \\
10-12 & $1.204\times 10^{10}$ & $\mathbf{1.231\times 10^{10}}$ \textbf{(+2.24\%)} \\
13-15 & $2.066\times 10^{9}$ & $\mathbf{2.114\times 10^{9}}$ \textbf{(+2.28\%)} \\
\bottomrule
\end{tabular}
}
\end{table}

\begin{table}[t]
\centering
\caption{
After Continual Pretraining on just 5B tokens (5\% of pretraining tokens), the average 0/5-shot accuracy on LLM benchmarks ($p$-value $<0.01$).
}
\begin{tabular}{c|cc}
\toprule
 & \textbf{Original BPE} & \textbf{LBPE} \\
\midrule
BoolQ & 59.86 & \underline{\textbf{60.26}} \\
HellaSwag & 41.12 & \underline{\textbf{41.54}} \\
OpenBookQA & 31.10 & \underline{\textbf{31.40}} \\
PIQA & 66.95 & \underline{\textbf{67.33}} \\
SIQA & 42.60 & \underline{\textbf{43.60}} \\
StoryCloze & 62.91 & \underline{\textbf{63.36}} \\
Winogrande & 53.47 & \underline{\textbf{53.63}} \\
\bottomrule
\end{tabular}
\label{continual_pretrain}
\end{table}

\subsubsection{Model}
\label{model}
Following Scaffold-BPE \cite{lian2024scaffold}, we train three LLMs with 468M, 1.2B, and 6.7B parameters, respectively. The architecture of 6.7B model is identical to LLaMA-7B \cite{touvron2023llama}.

\subsubsection{Training}
\label{Training}
Following previous works \cite{lian2024scaffold,su2024maskmoe,xiong2024temporal,su2024mile} and limited by our computation budget, all models are pretrained with 100B tokens. Note that the volume of corresponding text data contained in an equal amount of tokens is slightly different between the two tokenizers. Considering the commonly used criteria (i.e., the token amount) of computation budget in LLM training, we compare experiments in the setting of an equal amount of training tokens. In the Section \ref{EqualText}, we further analyze both tokenizers in the setting of an equal amount of training text volume.

\subsection{Experimental Results}
We incorporate 7 benchmarks recognized for evaluating LLMs, including BoolQ \cite{clark2019boolq}, HellaSwag \cite{zellers2019hellaswag}, OpenBookQA \cite{mihaylov2018can}, PIQA \cite{bisk2020piqa}, SIQA \cite{sap2019socialiqa}, StoryCloze \cite{mostafazadeh2016corpus}, and Winogrande \cite{sakaguchi2021winogrande}. We present the average 0/5-shot accuracy of all models.
As shown in Table \ref{Language Modeling Results}, we can observe that the LBPE consistently outperforms the original BPE with any model size. We conduct a $t$-test, and all metrics have $p$-values less than 0.01, indicating that the results are statistically significant.
Such results clearly demonstrate that our proposed LBPE is convincingly effective. We attribute it to that LBPE can encode text with more long tokens, which can help LLMs learn long tokens more thoroughly and mitigate the imbalanced-learning issue \cite{su2024mile}.

\begin{table}[t]
\centering
\caption{
At varying vocabulary sizes, the average 0/5-shot accuracy on LLM benchmarks ($p$-value $<0.01$).
}
\resizebox{\linewidth}{!}{
\begin{tabular}{c|cc|cc}
\toprule
& \multicolumn{2}{c|}{\textbf{64K}} & \multicolumn{2}{c}{\textbf{128K}} \\
& \textbf{Orig. BPE} & \textbf{LBPE} & \textbf{Orig. BPE} & \textbf{LBPE} \\
\midrule
BoolQ & 58.01 & \underline{\textbf{59.53}} & 56.67 & \underline{\textbf{58.36}} \\
HellaSwag & 41.82 & \underline{\textbf{42.28}} & 42.70 & \underline{\textbf{43.32}} \\
OpenBookQA & 30.90 & \underline{\textbf{31.40}} & 31.10 & \underline{\textbf{31.50}} \\
PIQA & \underline{\textbf{67.95}} & \underline{\textbf{67.95}} & 67.68 & \underline{\textbf{68.01}} \\
SIQA & 43.47 & \underline{\textbf{43.99}} & 43.83 & \underline{\textbf{44.93}} \\
StoryCloze & 64.19 & \underline{\textbf{64.54}} & 64.11 & \underline{\textbf{65.71}} \\
Winogrande & 53.67 & \underline{\textbf{54.78}} & 53.91 & \underline{\textbf{56.08}} \\
\bottomrule
\end{tabular}
}
\label{Vocab Size Results}
\end{table}

\subsection{Discussion}

\subsubsection{Token Length Distribution}
To demonstrate that LBPE can increase the frequency of longer tokens to mitigate the imbalanced learning issue, we derive the distribution of the token lengths in the Pile dataset tokenized by the original BPE and LBPE. As shown in Table \ref{Token Length Distribution}, the frequency of short tokens (i.e., the length of token less than 4 characters) decreases by 0.97\%
, while the frequency of tokens with a length greater than or equal to 4 characters consistently increases. Specifically, the frequency of tokens with lengths of 7-9, 10-12 and 13-15 characters can increase by 2.37\%, 2.24\%, and 2.28\%, respectively. Such an increase of long token frequency can help language models learn complex words more sufficiently.

\subsubsection{Never too Late to Use LBPE}
Currently, many LLMs have been trained on trillions of tokens, making it uneconomical to retrain existing LLMs from scratch using LBPE. And thus, we further verify whether it is feasible to switch from BPE to LBPE during a continual pretraining stage. As shown in Table \ref{continual_pretrain}, after training the 468M model with the original BPE for 100B tokens, switching to LBPE and continuing pretraining for just 5B tokens  (5\% of pretraining tokens) results in a substantial improvement in model performance compared to no switching (i.e., original BPE). Such a result demonstrates that LBPE has an immediate effect on improving model performance, and it's never too late to use LPBE.

\subsubsection{Various Vocabulary Sizes}
Depending on the size of the training corpus, the diversity of the languages, the size of the model, and the types of tasks, different vocabulary sizes are set in practice. Therefore, to validate the robustness of LBPE across various vocabulary sizes, in addition to the 32K vocabulary \cite{touvron2023llama}, we also trained two vocabularies sized at 64K \cite{baichuan7b,baichuan13b} and 128K \cite{yang2023baichuan}.
The experimental setup is identical to that of Section \ref{Experimental Setup}. 
As shown in Table \ref{Vocab Size Results}, LBPE consistently outperforms the original BPE across all vocabulary sizes, which indicates that the superiority of LBPE is not sensitive to vocabulary size. Its long-token-first tokenization mechanism can help LLMs mitigate the imbalanced-learning issue across different vocabulary sizes.

\begin{table}[t]
\centering
\caption{
At 300B training tokens, the average 0/5-shot accuracy on LLM benchmarks ($p$-value $<0.01$).
}
\begin{tabular}{c|cc}
\toprule
 & \textbf{Original BPE} & \textbf{LBPE} \\
\midrule
BoolQ & 58.21 & \underline{\textbf{61.48}} \\
HellaSwag & 45.10 & \underline{\textbf{46.26}} \\
OpenBookQA & 31.10 & \underline{\textbf{33.00}} \\
PIQA & 68.63 & \underline{\textbf{69.91}} \\
SIQA & 43.78 & \underline{\textbf{44.29}} \\
StoryCloze & 65.53 & \underline{\textbf{65.79}} \\
Winogrande & 53.67 & \underline{\textbf{55.64}} \\
\bottomrule
\end{tabular}
\label{Tokens Results}
\end{table}

\subsubsection{More Training Tokens}
According to the Scaling Law, the loss scales as a power-law with model size, dataset size, and the amount of training computation \cite{kaplan2020scaling}. To demonstrate the effectiveness of our LBPE with more training tokens, we continue training the 468M models up to 300B tokens \cite{zhang2022opt,biderman2023pythia}. 
As shown in Table \ref{Tokens Results}, LBPE consistently outperforms the original BPE at 300B training tokens, well indicating the effectiveness of LBPE in the era of increasingly large datasets for training LLMs.

\subsubsection{Higher Compression Rate}
Besides the performance of models, the compression rate for a given text corpus is a metric to measure the effectiveness of a tokenizer. A higher compression rate means that fewer tokens are required to represent the same corpus. As shown in Table \ref{Compression Rate}, LBPE intends to represent corpus with long tokens, thus can achieve a higher compression rate for a given vocabulary.

\begin{table}[t]
\centering
\caption{
Compression Rate (the average number of bytes per token) on the Pile dataset.
}
\begin{tabular}{c|ccccccc}
\toprule
 & \textbf{32K}  & \textbf{64K} & \textbf{128K}\\
\midrule
Original BPE & 3.4318 & 3.5777 & 3.6817 \\
LBPE & \underline{\textbf{3.4381}} & \underline{\textbf{3.5812}} & \underline{\textbf{3.6824}} \\
\bottomrule
\end{tabular}
\label{Compression Rate}
\end{table}

\begin{table}[t]
\centering
\caption{
At exactly 388 GiB training text, the average 0/5-shot accuracy on LLM benchmarks ($p$-value $<0.01$).
}
\begin{tabular}
{c|cccccccc}
\toprule
 & \textbf{Original BPE} & \textbf{LBPE} \\
\midrule
BoolQ & 58.72 & \underline{\textbf{60.88}} \\
HellaSwag & 40.84 & \underline{\textbf{41.62}} \\
OpenBookQA & 30.55 & \underline{\textbf{31.75}} \\
PIQA & 66.58 & \underline{\textbf{67.28}} \\
SIQA & 43.40 & \underline{\textbf{43.70}} \\
StoryCloze & 62.85 & \underline{\textbf{63.50}} \\
Winogrande & 53.07 & \underline{\textbf{54.79}} \\
\bottomrule
\end{tabular}
\label{Text Results}
\end{table}

\begin{table}[t]
\centering
\caption{
Combining LBPE with Scaffold-BPE, the average 0/5-shot accuracy on LLM benchmarks ($p$-value $<0.01$).
}
\resizebox{\linewidth}{!}{
\begin{tabular}
{c|ccc|ccccc}
\toprule
 & \textbf{Orig. BPE} & \textbf{LBPE} & \textbf{Scaffold-BPE} & \textbf{Scaffold-BPE + LPE} \\
\midrule
BoolQ & 58.64 & \underline{\textbf{60.89}} & 60.52 & 60.40 \\
HellaSwag & 40.78 & 41.61 & 41.68 & \underline{\textbf{42.17}} \\
OpenBookQA & 30.50 & 31.70 & \underline{\textbf{32.20}} & \underline{\textbf{32.20}} \\
PIQA & 66.57 & 67.08 & 68.69 & \underline{\textbf{69.15}} \\
SIQA & 43.40 & 43.71 & 44.09 & \underline{\textbf{44.27}} \\
StoryCloze & 62.77 & 63.52 & 63.04 & \underline{\textbf{64.64}} \\
Winogrande & 53.00 & 54.81 & 54.22 & \underline{\textbf{55.25}} \\
\bottomrule
\end{tabular}
}
\label{Scaffold Results}
\end{table}

\subsubsection{Experiments under Same Corpus Size}
\label{EqualText}
As mentioned in Section \ref{Training}, considering the commonly used criteria (i.e., the token amount) of computation budget in LLM training, experiments above are compared in the setting of an equal amount of training tokens. To eliminate the impact of different amounts of training text caused by different compression rates on results, we additionally train two 468M-parameter models on exactly 388 GiB training text ($\approx$ 100B tokens). As shown in Table \ref{Text Results}, LBPE consistently outperforms the original BPE, demonstrating that the effectiveness of LBPE is not merely obtained by allowing models to digest more data in the same computation budget. LBPE also alleviates the imbalanced learning issue, allowing models to learn all tokens more sufficiently and evenly, thus achieving improvements.

\subsubsection{Compatible with Other BPE Enhancements}
LBPE is orthogonal to and can be combined with existing enhancements to BPE, like Scaffold-BPE \cite{lian2024scaffold}. We replace the encoding algorithm of Scaffold-BPE with LBPE. As shown in Table \ref{Scaffold Results}, LBPE with Scaffold-BPE achieves further improvements, well indicating the compatibility of LBPE.

\section{Conclusions}
We propose LBPE, aimed at addressing the learning imbalance issue caused by vastly different frequencies of short/long tokens in LLMs pretraining. By prioritizing longer tokens during the encoding process, LBPE effectively enriches their occurrences.
Experimental results demonstrate that LBPE consistently outperforms the original BPE. Particularly, LBPE is useful in the continual pretraining stage and it's never too late to use LBPE. Moreover, LBPE is orthogonal to existing modifications on BPE, and can be combined with them to achieve further improvements.



\bibliographystyle{IEEEtran}
\bibliography{IEEEabrv,mybibfile}

\end{document}